# Knowledge Engineering Within a Generalized Bayesian Framework

Stephen W. Barth and Steven W. Norton*
PAR Government Systems Corporation
220 Seneca Turnpike
New Hartford, NY 13413
(315) 738-0600
20 May 1986

## Abstract

During the ongoing debate over the representation of uncertainty In Artificial Intelligence, Cheeseman, Lemmer, Pearl, and others have argued that probability theory, and in particular the Bayesian theory, should be used as the basis for the inference mechansims of Expert Systems dealing with uncertainty. In order to pursue the issue in a practical setting, sophisticated tools for knowledge engineering are needed that allow flexible and understandable Interaction with the underlying knowledge representation schemes. This paper describes a Generalized Bayesian framework for building expert systems which function in uncertain domains, using algorithms proposed by Lemmer. It is neither rule-based nor frame-based, and requires a new system of knowledge engineering tools. The framework we describe provides a knowledge-based system architecture with an Inference engine, explanation capability, and a unique aid for building consistent knowledge bases.

## 1. Introduction

Recent arguments have been made for the use of Bayesian inferencing techniques for expert systems dealing with uncertainty [Charniak, 1983; Cheeseman, 1986; Pearl, 1982; Yeh, 1983]. Some expert system (ES) applications involving reasoning under uncertainty may best be handled by mechanisms that are as purely Bayesian as possible. This paper describes a Generalized Bayesian framework for constructing expert systems using networks of probability distributions, based on Lemmer's approach to Generalized Bayesian Inference (GBI) [Lemmer, 1976; Lemmer, 1983; Lemmer and Barth, 1982].

The knowledge representation scheme for our GBI framework consists of a network of intersecting sets of events, and their associated probability distributions. The GBI Framework contains the key ingredients of a classic knowledge-based system architecture: an inference engine with an explanation capability, and a tool for creating, editing, and maintaining consistency in the knowledge base. These individual components, however, are very different from those of rule-based probabilistic systems such as PROSPECTOR [Duda, Hart, et al., 1979]. The purpose of this paper is to describe the GBI framework for building expert systems, its tools for estimating consistent prior probability distributions (the CMD Estimation Aid), and its facilities supporting the utilization and incremental development of knowledge bases, the Updating Mechanism and Explanation Facility. The description of the Explanation Facility will be ommitted here, since it is described in detail in [Norton, 1986] in this volume.

The system described in this paper was developed for application to a problem of real-time discrimination of object types from sensor information. The original approach to this problem involved using a Markov model with a sequence of discriminant tests to classify objects. This approach had several difficulties. Order dependancies among the tests made it hard to tune the system for accurate classifications. Heuristic estimates of probabilities incorporated into state transition matrices caused instabilities in the decision process. In general, the original approach lacked a framework for understanding the results from system test and evaluation and using those results for improving system performance. A knowledge-based approach to the problem was taken to alleviate these problems. The Generalized Bayesian Framework was developed to provide the tool for knowledge-based handling of the rich environment of probabilistic information available for the problem.

Unfortunately the details of the problem domain are classified. Therefore, in this paper, we will illustrate knowledge representation in the system with a small knowledge base for predicting the weather in

---

* Steven W. Norton is now employed at Siemens RTL, 105 College Rd. East, Princeton, NJ 08540

Upstate New York; in particular, providing expert estimates of the probability of precipitation. This toy problem example has a similar diagnostic character to the real problem: the goal is to infer the correct identification or prediction from observed characteristics or measurable quantities. The advice provided by the system, or an expert, for both problem domains, always takes the form of a probabilistic recommendation, because of the uncertainty of the data and its interpretation.

## 2. Knowledge Engineering within the Generalized Bayesian Framework

GBI is a tool for building expert systems to solve Bayesian hierarchical inferencing problems. The initial phases of knowledge engineering for an expert system using this framework proceed as in any other hierarchical inferencing problem [Hayes-Roth, Waterman, and Lenat 1983]. Together with one or more domain experts, the knowledge engineer considers the expert thought process used in solving the problem at hand. A series of stages is sketched out which takes observable evidence, relates it to intermediate hypotheses, and uses these intermediate hypotheses to determine the validity of one or more goal variables. The problem variables (evidence, hypothesis, and goal variables) represent assertions of interest. Sets of these variables form the nodes in the network structure of a GBI knowledge base.

In a typical rule-based expert system framework, the knowledge engineer, in cooperation with one or more domain experts, specifies the relationships between the problem variables using rules. In GBI, the knowledge engineer groups together those variables for which some probabilistic constraints (correlation information) are known. These structures are called Local Event Groups by Konolige in an appendix to [Duda, Hart, et al, 1979]. A LEG Network (LEG Net) is a collection of LEGs, some of which share common variables. The LEG Net structure arises naturally from the hierarchical problem organization.

The aim of our sample knowledge base is to combine information from many sources to match the weather predicting capability of local experts in Upstate New York. An initial LEG Net for this knowledge base is shown in Figure 1. Each LEG is indicated by a round edged box labeled with a name in bold type. The names of the variables in the LEG are listed inside the box. Square shaded boxes show the variables in LEG intersections, and are attached to lines connecting the intersecting LEGs.

In our example the goal variables are Snow-Tomorrow, Rain-Tomorrow, and No-Precip-Tomorrow, in the Kind-of-Precip LEG, representing the events or assertions "it will rain tomorrow," "it will snow tomorrow," and "there will be no precipitation tomorrow." The advice to the user will be whatever values are attained by their posterior probabilities. The hypothesis variables Folk-Precip and Others-Precip represent the assertions that folk signs and other-agencies should be considered in predicting precipitation for tomorrow. Examples of evidence variables are Moon-Haze and Bunions-Ache in the Folk-Precip LEG. Bunions-Ache will be true if Grandma has complained about her feet lately, and Moon-Haze will be true if haze appears around the moon. Other evidence variables in the example are mutually exclusive. In Figure 1 mutually exclusive events are shown boxed within their LEG. The evidence variables in this example represent events that either occur or do not occur and hence have a posterior probability of either 0.0 or 1.0. Such variables are called Binary Evidence Variables or BEVs.

If all the problem variables were grouped together, we could, in principle, establish a single underlying probability distribution. In practice that is quite infeasible since the storage requirements for probability distributions are exponential ($2^n$) in terms of the number of variables covered. Even for the simple weather example of Figure 1, over 8 million probabilities may be required for the underlying distribution.

Lemmer's GBI approach avoids this problem by using only Component Marginal Distributions (CMDs) for each LEG. [Lemmer, 1983]. Each LEG in a LEG Net is associated with a CMD which covers just those variables in the LEG. Taken together, the CMDs for a LEG Net are an approximation to the underlying distribution. It is an approximation in the sense that no n-variate constraints are specified. If the largest LEG contains only $m$ variables then the LEG Net can contain no constraints of order greater than $m$. This approximation is reasonable since by breaking up the event space into LEGs we are specifying which constraints are important, and which are not. The size requirement for the network is the total of the sizes of the CMDs. The CMDs which are to be specified for the network in Figure 1 will have a total of 368 elements versus more than 8 million in the underlying distribution.

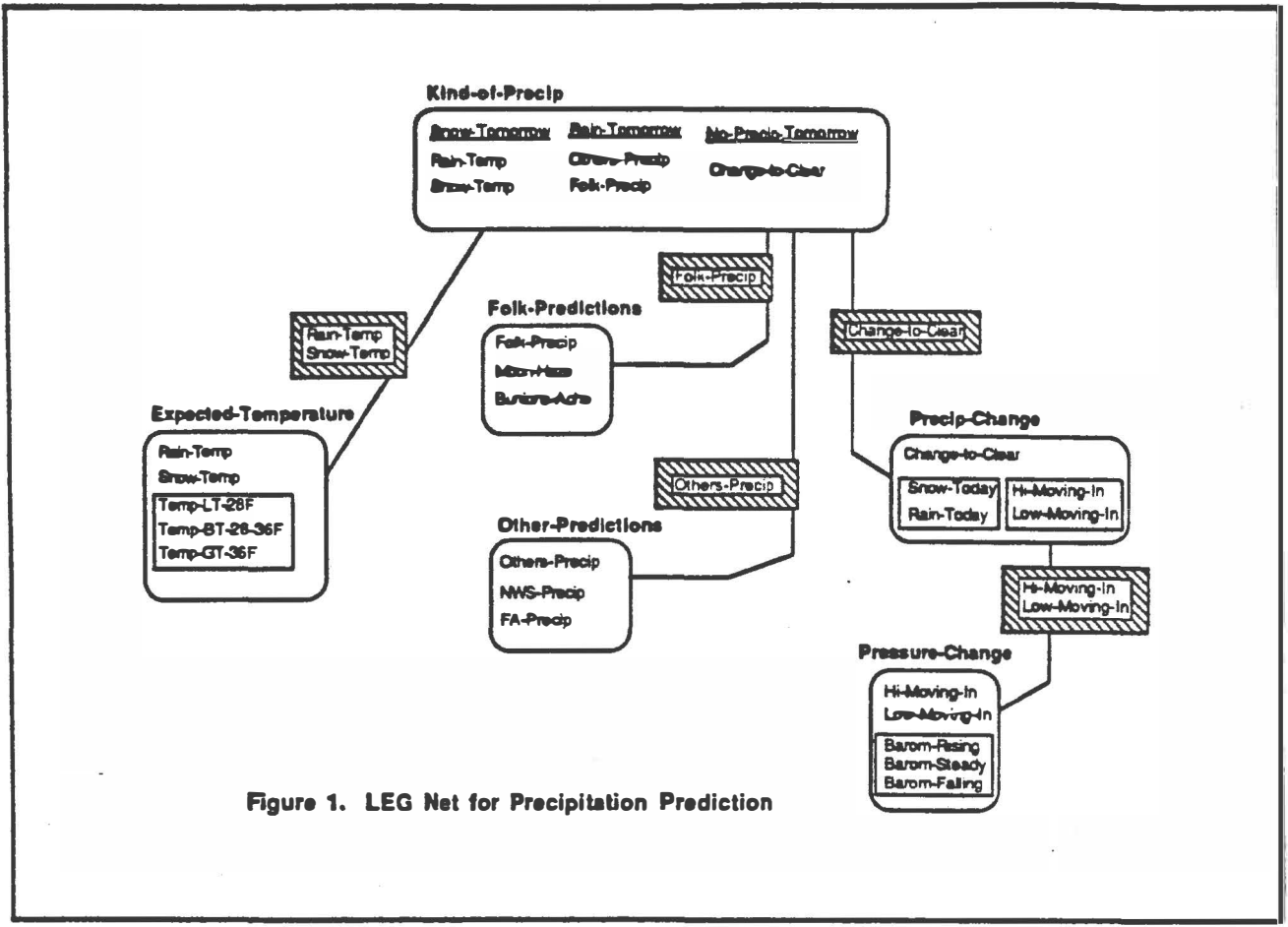

Figure 1. LEG Net for Precipitation Prediction

The Lemmer approach requires $2^m - 1$ constraints to fully specify an m-variate CMD. However, it allows the system to make use of information about mutually exclusive or implied events which may cause many of the constraints to be set to zero. The CMD Estimation Aid exploits this capability, so that in practice the knowledge engineer has many fewer probabilities to determine or estimate than the product of the number of LEGs and ($2^m - 1$). The number of constraints to be specified can be reduced still further by accepting minimum information default values for the higher order constraints, which are automatically generated. The examples in the next section illustrate these features of the system.

## 2.1 Specifying LEGs and CMDs

The CMD Estimation Aid of the GBI framework provides an interactive, menu-driven interface for defining events, LEGs, LEG Nets, and estimating CMDs. LEG Intersections, through which probability is propagated, are managed automatically. In the following paragraphs we describe knowledge engineering within the GBI framework by walking through the features of the CMD Estimation Aid and illustrating the thought processes used during estimation in terms of the example in Figure 1.

One body of evidence weighed by our local expert is that of precipitation predictions from other sources. Our example considers two: the probability of precipitation predicted by the National Weather Service, NWS--Precip, and that predicted by the Farmers Almanac, FA--Precip. The heuristic that our local expert uses is that when either one of these sources predicts precipitation it increases the chances of rain or snow tomorrow. The hypothesis variable Others-Precip is introduced to indicate the probability of precipitation predicted from either of these two sources, and we form the LEG Other-Predictions to handle it and the two evidence variables. The constraints on the distribution for Other-Predictions are that if either or both of the variables NWS--Precip and FA--Precip are certain, then Others-Precip is certain. In addition, let us say we know, from statistical information, that the National Weather Service predicts precipitation 55% of the time, that the Farmers Almanac predicts precipitation 45% of the time, and that the two sources agree on the prediction 35% of the time.

Using the CMD Estimation aid requires the knowledge engineer to specify the constraints on a LEG in a canonical order, to support efficiency in the algorithm for constructing the distribution; however, the knowledge engineer has the freedom to select the style of constraints by choosing the order of variables for consideration and the form of the constraint that is most easily understood or calculated, either a joint or conditional probability. (The details of the CMD estimation algorithms are described in [Lemmer, 1976] and [Lemmer and Norton, 1986].) In the LEG Other-Predictions we chose the constraints with a diagnostic "style" in mind; since we wanted to specify the conditional probabilities of the hypothesis variable given the evidence variables as symptoms. To accomplish this we proceed by specifying the marginal probabilities for FA-Precip and NWS-Precip first, 0.45 and 0.55, respectively. The system prompts next for either the conditional of FA-Precip given NWS-Precip, or the marginal joint probability of FA-Precip and NWS-Precip. The latter we know to be 0.35. Now, we must specify the marginal probability for the hypothesis variable Others-Precip. A simple calculation with a Venn diagram shows that it must be 0.65. From here on out the system prompts us for constraints of either Others-Precip conditioned on the evidence variables, or the related joint probabilities. The former is easiest to specify, since P(Others-Precip | FA-Precip), P(Others-Precip| NWS-Precip), and P(Others-Precip | FA-Precip, NWS-Precip) are all 1.0. Figure 2 shows the constraints in the canonical order and the resulting distribution for the Other-Predictions LEG. The bit patterns in the left column under CMD indicate the joint events in the distribution and the right column holds their probabilities.

| | | |
|---|---|---|
| Bit 0 | F | FA-Precip |
| Bit 1 | N | NWS-Precip |
| Bit 2 | P | Others-Precip |

| Constraints | | CMD | |
|---|---|---|---|
| | | 000 | 0.35 |
| Pr(F) | 0.45 | 001 | 0.00 |
| Pr(N) | 0.55 | 010 | 0.00 |
| Pr(N&F) | 0.35 | 011 | 0.00 |
| Pr(P) | 0.65 | 100 | 0.00 |
| Pr(P|F) | 1.00 | 101 | 0.10 |
| Pr(P|N) | 1.00 | 110 | 0.20 |
| Pr(P|N&F) | 1.00 | 111 | 0.35 |

**Figure 2  Constraint Set for Other-Predictions**

As another example, consider the LEG Folk-Predictions. Here, the knowledge engineer wants to weigh together the results from two evidence variables, Bunions-Ache, and Moon-Haze. The former we estimate to occur about 45% of the time, the latter about 65%, and that both events happen together about 30% of the time. The hypothesis variable Folk-Precip represents the belief we have that the folk signs indicate rain, which we estimate at 0.55; i.e. they are slightly predisposed to predict rain. If there's haze around the moon, however, we're 60% sure that a folksy prediction is being made. If Grandma is complaining about her feet, we're 80% sure. And if both of these events occur then it's 99% certain that folk signs are telling us that precipitation will occur. The constraints just described are summarized in Figure 3. The effect of knowing that the folk signs indicate rain on our actual prediction is taken account of in the constraints on the goal leg between Folk-Precip and the other variables.

```
           Bit 0  M   Moon-Haze
           Bit 1  G   Bunions-Ache
           Bit 2  P   Folk-Precip

    Constraints                      CMD
                              000   0.1255
    Pr(M)       0.65          001   0.2570
    Pr(G)       0.45          010   0.0645
    Pr(G&M)     0.30          011   0.0030
    Pr(P)       0.55          100   0.0745
    Pr(P|M)     0.60          101   0.0930
    Pr(P|G)     0.85          110   0.0855
    Pr(P|G&M)   0.99          111   0.2970
```

**Figure 3  Constraint Set  for Folk-Predictions**

### 2.2 Expediting CMD Estimation

  The examples above show how a simple probabilistic OR relationship and weighing of beliefs can be specified using the CMD Estimation Aid. Generally, however, the knowledge engineer needs to make use of shortcut facilities that the CMD estimation aid provides. These are: allowing minimum information values as defaults for constraint probabilities, and taking advantage of "forbidden" and "cutoff" relationships among events. Whenever the CMD Estimation Aid prompts the user to specify a constraint it offers a range of values possible for the constraint that is consistent with the previously specified constraints. Within this range the value offered as a default for the constraint is the probability that satisfies a minimum information condition on the distribution under construction [Lemmer and Norton 1986]. In the example above the default value for the constraint P(FA-Precip & NWS-Precip) was 0.2475, which represents independence of the two variables. For small LEGs of only three or four variables, or larger ones that are highly constrained by cutoff or forbidden relationships, such default values are not so important except as a convenient function of the user interface. In other cases the ability to take default values for constraints that are needed to determine the distribution, but for which values are not known or cannot be estimated easily, is a critical feature. For example in the goal LEG Kind-Of-Precip, the user may not feel comfortable estimating constraints on more than four variables, such as P(Rain-Temp, Snow-Temp, Others-Precip, Folk-Precip), or any conditional form of these four variables, though it is not desirable to decompose Kind-Of-Precip into smaller LEGs. The mechanism for default values allows the user to ignore constraints above a certain order, or specify only those for which some justification is available, taking defaults for the rest.

  The CMD Estimation Aid also allows the user to specify "forbidden" and "cutoff" relationships between variables that automatically limit the number of constraints required to specify the distribution. Forbidden relationships indicate that the occurrence of two variables in a joint event is impossible, as is the case for mutually exclusive  events. In the LEG Expected-Temperature the three evidence variables Temp-LT-28F, Temp-BT-28-36F, and Temp-GT-36F are mutually exclusive (and collectively exhaustive) and so are given the forbidden relationship.  As a result, the knowledge engineer needs to specify only 12 constraints on joint events, instead of 26, and, if defaults are taken for constraints on more than two variables, then only six constraint  probabilities need to be specified.

  Cutoff relationships between variables indicate that one variable cannot occur in a joint event unless another does, in an implication relation. For example, in the Other-Predictions LEG, we might have specified that NWS--Precip only occurs if FA--Precip occurs. This means that joint events in which NWS--Precip occur, but FA--Precip does not occur, have probability 0.0. Thus the number of constraints required to specify the distribution is reduced.

11

The knowledge engineer can make use of all of these mechanisms to reflect the expertise being modeled, and to reduce the amount of information needed to define the knowledge base. As might be expected, our experience has been that cutoff and forbidden relationships nearly always exist among evidence variables but are harder to find in LEGs where hypothesis variables have been introduced to represent a concept or summarize an effect. In either case, but especially the latter, the default values mechanism makes it easy to specify what low order constraints are known for a distribution, while letting the system make reasonable assumptions for the higher order constraints that the expert or knowledge engineer may be unable to specify.

## 3. Generalized Bayesian Inference

Once prior CMDs for each LEG have been estimated by the knowledge engineer using the CMD Estimation Aid, posterior CMDs are calculated by the GBI Updating mechanism as evidence is gathered; i.e. as the posterior probabilities of the evidence variables are determined from observable data. Summing out the marginal probability of a LEG variable from a posterior CMD provides the advice about the occurrence of the event represented by that variable, based on the evidence gathered so far. As more evidence is gathered, a new posterior CMD may be calculated for a LEG, and new marginal probabilities determined for the LEG variables. When the posterior probabilities for all evidence variables have been determined the marginal probabilities of the goal variables represent the final advice of the system.

The GBI Updating rule has been formally described in [Lemmer, 1983] and [Lemmer and Barth, 1982]. Inferences are propagated through the CMDs over LEG intersections. If $A$ and $B$ are two intersecting LEGs, the formula for updating the CMD of $A$ after a new posterior CMD for $B$ has been determined is:

$$Pr'(A,a) = Pr(A,a) \cdot \frac{\sum_{\substack{b \subseteq B \\ a \cap I = b \cap I}} Pr'(B,b)}{\sum_{\substack{b \subseteq B \\ a \cap I = b \cap I}} Pr(B,b)} \quad \text{where } I = A \cap B \qquad (1)$$

where $a$ is a subset of the variables in $A$, and $P(A,a)$ represents the prior probability for the joint event in which the variables of $a$ occur and the other variables of $A$ do not. (As $a$ ranges over the subsets of $A$, $P(A,a)$ ranges over the prior probabilities of the joint events in $A$'s CMD). Similarly, $P'(A,a)$ represents the posterior probability of the joint event in which the variables of $a$ occur and the other variables of $A$ do not.

It is important to realize that $P(A,a)$ represents probabilities in the most recent posterior that was determined for $A$. Initially $P(A,a)$ has the value that was assigned during the knowledge engineering phase with the CMD Estimation Aid, but as evidence is collected and udating proceeds, several new CMDs may be calculated for $A$. Each time a new CMD needs to be calculated the $P(A,a)$ values in the formula are the ones from the last update to the CMD, not the *a priori* distrbution.

Viewed as a sequential process, the updating formula means that for each joint event $e$ in $A$ ($P(e) = P(A,a)$ for some $a$) we

1. Determine the joint event i in the LEG intersection of $A$ and $B$ which is a component of $e$.
2. Sum out from the previous CMD of $B$ to obtain the prior probability of i. This is the denominator of the fraction given in the formula.
3. Sum out from the new CMD of $B$ to obtain the new posterior marginal probability of i. This is the numerator of the fraction in the formula.
4. Multiply P(e) by the quotient of the values obtained in 2 and 3 above to obtain its posterior P'(e).

As an example of the Updating mechanism consider an update of the CMD of the goal LEG Kind-of-Precip from the evidence LEG Folk-Predictions in Figure 1. Figure 4 shows the transformations of the

12

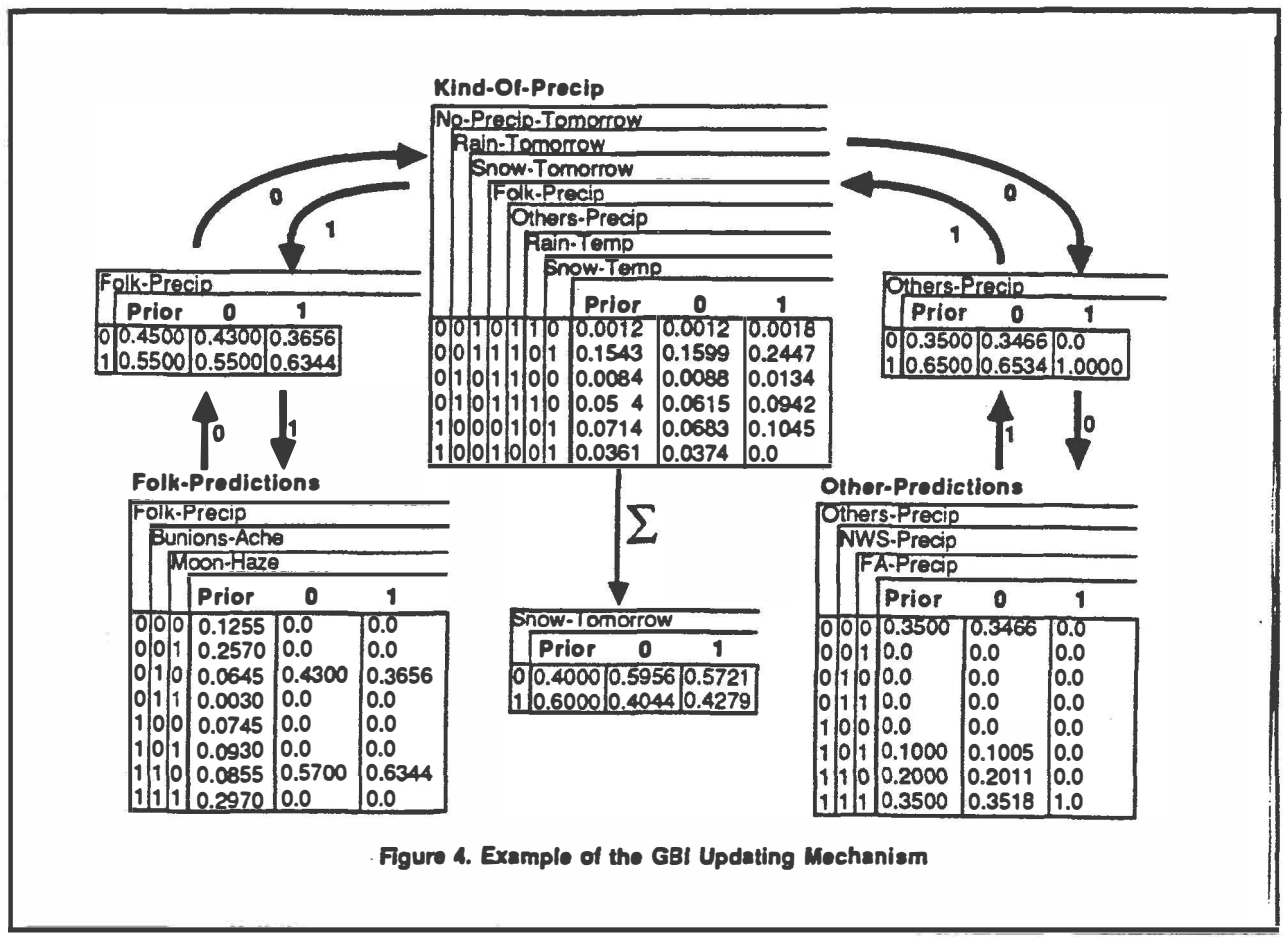
Figure 4. Example of the GBI Updating Mechanism

CMDs for part of the Kind-of-Precip LEG, and the entire Folk-Predictions and Other-Precip- Predictions LEGs, for the situation in which the events Bunions-Ache, NWS--Precip, and FA--Precip occur.

After it is determined (through querying the user or accessing a data base) that the event Bunions-Ache occurs and that Moon-Haze does not occur, the posterior distribution of Folk-Predictions is calculated by applying the GBI Updating rule with the marginal distribution of the two evidence events as the updating distribution. In this instance, the GBI Updating rule is equivalent to Bayes' Rule applied to the CMD of Folk-Predictions, since the conditioning events are certain. [Lemmer 1983]. In the next step, the new CMD for the intersection of Folk-Predictions and Kind-of-Precip is calculated by summing out over the CMD of Folk-Predictions. Now, the CMD of Kind-of-Precip is calculated using the GBI Updating rule. For example, the prior probability of the joint event that Snow-Tomorrow, Folk-Precip, Others-Precip, and Snow-Temp occurred, but No-Precip-Tomorrow, Rain-Tomorrow, and Rain-Temp did not occur, in the Kind-of-Precip LEG, is multiplied by the quotient of the posterior and prior probabilities for the event Folk-Precip in the LEG intersection. (0.1599 = 0.1543 * 0.5700 / 0.5500). Next the CMD of Other-Predictions is updated using the CMD for the LEG intersection between Kind- of-Precip and Others-Predictions. The posterior CMD for the LEG intersection is calculated by summing out from the new posterior CMD for Kind-of-Precip to obtain the probabilities of Others-Precip and not Others-Precip. The GBI Updating rule is then used to calculate the new posterior CMD for the Other-Predictions LEG. Posterior CMDs for the other LEGs would also be calculated from the effects of the update to Folk-Predictions so that consistent CMDs are maintained throughout the network after the completion of each update.

The fact that propagation of probability occurs throughout the LEG Net is an important difference between the the GBI Updating rule and the mechanisms used in systems like MYCIN [Buchanan and Shortliffe, 1984] and PROSPECTOR. In PROSPECTOR for example, probability is only propagated from evidence to goal variables, and the prior probabilities for unexamined evidence remain unmodified. In the GBI mechanism consistent probability distributions are maintained: summing out the marginal probability for a

13

variable in a LEG intersection from either CMD that it belongs to yields the same value. A practical advantage of the GBI mechanism is that it allows the system to look for the evidence that is next most or least likely to occur, determined from the marginal probabilities of evidence variables summed out from their most recent posterior CMD. Whether such evidence does or does not in fact occur in a given situtation can be used to control the inference mechanism; e.g. by computing expected costs and benefits of evidence tests.

## 4. Generalized Bayesian Explanation

As with any Expert System knowledge representation scheme, the GBI representation offers a means of explaining inferences which provides a knowledge engineer with the information needed to tune the knowledge base. A trace of the inference mechanism is available during a consultation session, when only partial results are available, or after the system has exhausted all evidence, as a post-mortem explanation. The GBI Explanation Mechanism Is described in "An Explanation Mechanism for Bayesian Inferencing Systems" [Norton 1986] appearing elsewhere in these proceedings.

## 5. Future Work

Although our GBI framework has been successfully applied to a real problem other than the example discussed here, much work remains to develop an industrial strength Expert System building tool. The issues we are currently concerned with are:

1. Mining CMDs for richer explanations of Updating Mechanism results. Much more useful information on expected and unexpected relationships between variables can be presented to the knowledge engineer to aid in tuning the knowledge base. Multi-variate relationships and the effects of initial prior constraints can be easily obtained. A better man-machine interface can be developed with graphics for use of the system in consultation mode.

2. Allowing distributions over evidence variables as inputs during Updating. Uncertain marginal probabilities for evidence variables, or, in cases where all possible data has been collected, for joint evidence events, define posterior CMDs for evidence LEGs other than those currently allowed in which some joint event must have unit probability.

3. Exploiting parallelism. The GBI framework has proven quite efficient for the relatively small problems to which it has been applied. However, the GBI Updating rule has characteristics which would allow fast implementations for larger knowledge bases.

4. Providing more CMD Estimation Aid facilities. Ways of easily backtracking through the canonical constraints and and other estimation shortcuts can be provided. A graphic presentation of the low order constraints, and a separate explanation mechanism for the CMD estimation process are also desirable.

## 6. Summary

Our GBI framework provides a knowledge-based system architecture for a Generalized Bayesian knowledge representation and inferencing technique. The CMD Estimation Aid provides the needed knowledge engineering tool for estimating distributions of prior probabilities over sets of events. The Updating Mechanism provides an inference engine for using the knowledge base to obtain advice in the form of marginal probabilities associated with goal variables. An explanation facility provides feedback to allow the knowledge engineer to refine the knowledge base.

Although the current system is in the embryonic stage, it has been successfully applied to develop a prototype knowledge base for discriminating types of objects from real time sensor data. With further development we feel that the GBI framework will allow knowledge engineers and application users to take advantage of the richness of the Bayesian representation for uncertain reasoning.

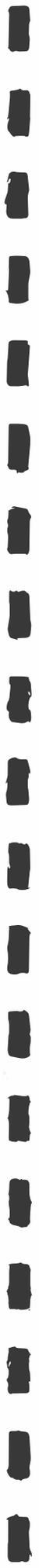